\documentclass[sigconf]{acmart}
\AtBeginDocument{%
  }

\setcopyright{acmlicensed}
\copyrightyear{2018}
\acmYear{2018}
\acmDOI{XXXXXXX.XXXXXXX}
\acmConference[Conference acronym 'XX]{Make sure to enter the correct
  conference title from your rights confirmation email}{June 03--05,
  2018}{Woodstock, NY}
\acmISBN{978-1-4503-XXXX-X/2018/06}

\usepackage{subfigure}
\usepackage{multirow}
\usepackage{enumitem}
\usepackage{newfloat}
\usepackage{listings}
\usepackage{tcolorbox}
\DeclareCaptionStyle{ruled}{labelfont=normalfont,labelsep=colon,strut=off} 
\floatstyle{ruled}
\newfloat{listing}{tb}{lst}{}
\floatname{listing}{Listing}
\begin{document}

\title{Label Forensics: Interpreting Hard Labels in Black-Box Text Classifier}

\author{Mengyao Du}
\email{dumengyao@nudt.edu.cn}
\orcid{0000000295602869}
\affiliation{%
  \institution{National University of Defense Technology}
  \state{Changsha}
  \country{China}
  \postcode{410005}
}

\author{Gang Yang}
\email{ygang@u.nus.edu}
\affiliation{%
  \institution{National University of Singapore}
  \country{Singapore}
}

\author{Han Fang}
\email{fanghan@nus.edu.sg}
\affiliation{%
  \institution{National University of Singapore}
  \country{Singapore}}

\author{Quanjun Yin}
\email{yin_quanjun@163.com}
\affiliation{%
  \institution{National University of Defense Technology}
  \city{Changsha}
  \country{China}}

\author{Ee-Chien Chang}
\email{	changec@comp.nus.edu.sg}
\affiliation{%
 \institution{National University of Singapore}
 \country{Singapore}}

\renewcommand{\shortauthors}{Trovato et al.}

\begin{abstract}

The widespread adoption of natural language processing techniques has led to an unprecedented growth of text classifiers across the modern web. Yet many of these models circulate with their internal semantics undocumented or even intentionally withheld. Such opaque classifiers, which may expose only hard-label outputs, can operate in unregulated web environments or be repurposed for unknown intents, raising legitimate forensic and auditing concerns. In this paper, we position ourselves as investigators and work to infer the semantic concept each label encodes in an undocumented black-box classifier.

Specifically, we introduce label forensics, a black-box framework that reconstructs a label’s semantic meaning. Concretely, we represent a label by a sentence embedding distribution from which any sample reliably reflects the concept the classifier has implicitly learned for that label. We believe this distribution should maintain two key properties: \textbf{\textit{precise}}, with samples consistently classified into the target label, and \textbf{\textit{general}}, covering the label’s broad semantic space. To realize this, we design a semantic neighborhood sampler and an iterative optimization procedure to select representative seed sentences that jointly maximize label consistency and distributional coverage. The final output, an optimized seed sentence set combined with the sampler, constitutes the empirical distribution representing the label’s semantics. Experiments on multiple black-box classifiers achieves an average label consistency of around 92.24\%, demonstrating that the embedding regions accurately capture each classifier’s label semantics. We further validate our framework on an undocumented HuggingFace classifier, with the resulting analysis also presented in this paper, enabling fine-grained label interpretation and supporting responsible AI auditing.


\end{abstract}



\received{20 February 2007}
\received[revised]{12 March 2009}
\received[accepted]{5 June 2009}

\maketitle

\section{Introduction}

Recent advances in NLP (Natural Language Processing) have led to the emergence of a wide range of powerful text classifiers~\cite{nlp1,nlp2,nlp3}. Many of these models are deployed on modern web platforms for tasks such as sentiment analysis~\cite{sentiment2}, spam filtering~\cite{spam2}, and toxicity detection~\cite{toxic2}. These models have greatly facilitated real-world applications by enabling automated content moderation~\cite{contentmoderation}, personalized recommendations~\cite{wang2018happiness}, and safety monitoring~\cite{safetymonitoring}, and are increasingly accessed through lightweight mechanisms such as inference interfaces or cloud-based deployments to achieve inference without maintaining local infrastructure~\cite{huggingface_inference_api,google_perspective_api,deng2022tee,dhar2025guardaintee}. 

\begin{figure}[t]
\centering
\includegraphics[width=0.95\columnwidth]{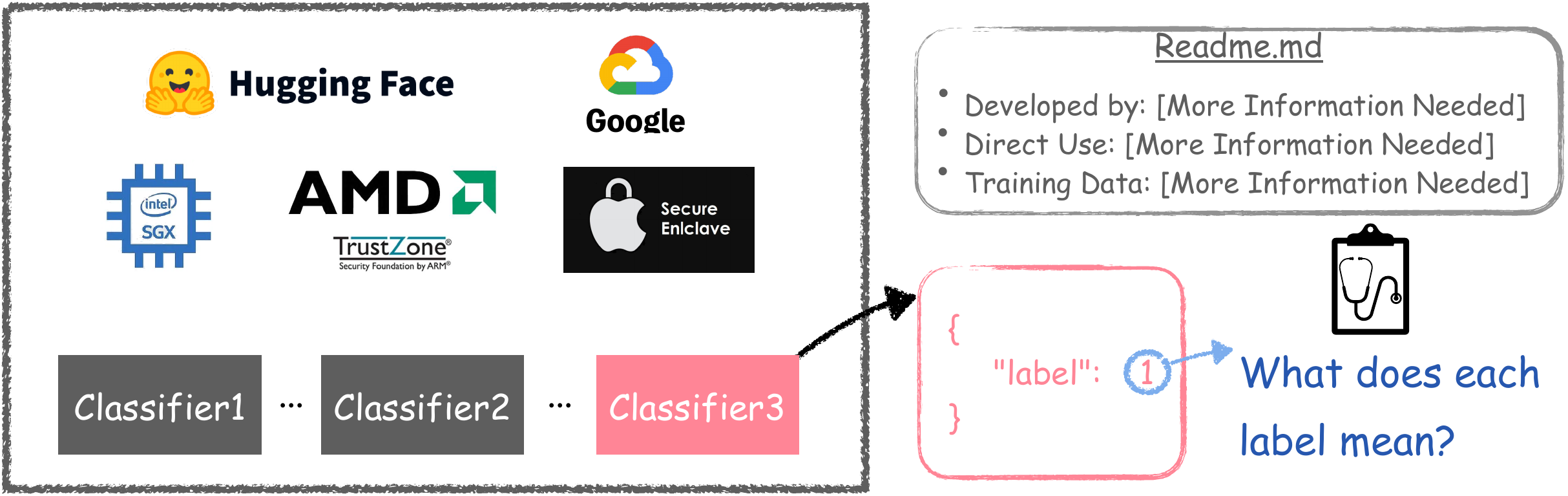} 
\caption{Application scenario of Label Forensics. Web-based text classifiers often return only JSON hard labels, motivating the need to investigate what these labels actually represent.}
\label{fig1}
\end{figure}

While many production-grade classifiers are well-documented and designed for legitimate use, a substantial number of models circulating on the web provide no documentation and offer little visibility into their intended purpose. These opaque classifiers often reveal only a hard-label output, as illustrated in Figure~\ref{fig1}, without exposing the semantics behind their predictions. Such lack of transparency creates risks in web environments where classifier outputs increasingly influence online interactions, civic engagement, and content governance. Such lack of transparency raises concerns about hidden biases, unintended misuse, or unknown decision logic. From the perspective of safe and responsible AI, this motivates a need for systematic forensic analysis capable of recovering the semantic meaning behind each output label, thereby supporting safe, interpretable, and accountable operation of text classification systems in real-world settings.

In this work, we take the role of an investigator, aiming to understand what a black-box text classifier is truly doing. Given only a text classifier, without documentation, label descriptions, or access to internal signals, we seek to recover the semantic distribution associated with each output label. Concretely, for every label, we aim to characterize the set of natural language sentences that the classifier consistently associates with that label, forming an empirical embedding distribution that reflects its underlying semantic concept. 


Achieving this goal is far from straightforward, especially under realistic black‑box conditions where the classifier returns only a one‑hot label for any input. From a distributional perspective, two key challenges arise. First, the model produces a label for every query, so random probing or isolated triggers offer little insight into what truly defines a class. Second, recovering a label’s semantic distribution requires identifying a set of representative sentences that collectively capture its conceptual meaning, rather than just finding individual inputs that elicit the label. This goal is fundamentally different from prior black‑box analysis methods, such as membership inference~\cite{he2025towardsmembership1,meeus2025sokmembership2,huang2025dfmia}, model inversion~\cite{morris2024promptinversion,lin2024inversion}, and training data extraction~\cite{icmlCarlini,yu2023bagdataextraction}, which focus on leaking or reconstructing specific training samples. Those techniques operate at the instance level, whereas our task demands distribution‑level semantic recovery, which demands methods that can capture broader and more systematic semantic patterns. Filling this gap is crucial for understanding what a classifier’s labels represent and for conducting trustworthy forensic analysis.

To address this challenge, we propose \textbf{Label Forensics}, a black‑box framework for reconstructing the semantic distribution of each label. We first define this distribution as a semantic region described by the set of natural language sentences that reflect the meaning of the label, along with the broader semantic space associated with them. This semantic region serves as an empirical characterization of the label’s underlying concept. We consider a good distribution to be precise, with samples reliably classified into the target label, and general, covering the broad semantic space of that label. These two properties ensure that the recovered distribution not only mirrors the classifier’s labeling behavior but also provides a faithful and interpretable representation of the underlying concept.


Our approach combines a broad candidate generator, a semantic neighborhood sampler, and an iterative search procedure to reconstruct the distribution of each label. We begin by assembling a diverse pool of label seed sentence sets intended to cover a wide range of linguistic expressions. The sampler, instantiated as a prefix-tuned encoder–decoder model which we train within our framework, generates local semantic variations around a seed sentence while preserving label consistency. The iterative search then refines a set of representative seeds by maximizing the objective, which jointly measures precision and coverage. Starting from an initial seed, the search expands neighborhoods, filters label‑consistent candidates, and incrementally builds an optimized seed set. The resulting seeds, together with the sampler, constitute the empirical distribution for each label, enabling faithful, interpretable, and distribution‑level forensics in black‑box settings. Experiments on real‑world text classifiers, including emotion, sentiment, spam, and jailbreak detection, demonstrate that label forensics yields precise and general semantic summaries, supporting scalable model auditing under opaque deployment. All code and experiments are released at an anonymized repository \footnote{\url{https://anonymous.4open.science/r/Label_Forensic-EEED}}.

\textbf{Contributions}:
\begin{itemize}
    \item We highlight the importance of black‑box text classifier forensics, which aims to recover the semantic meaning behind each output label under hard‑label and minimal‑access settings, forming a principled foundation for auditing opaque classifiers.
    \item We propose label forensics, a general-purpose framework that reconstructs label semantics as empirical distributions over natural language sentences. The method combines a prefix-tuned semantic neighborhood sampler with a geometric optimization objective to ensure that the resulting distributions are both \textit{precise} and \textit{general}, enabling faithful and interpretable semantic recovery.
    \item We evaluate our approach across five text classifiers spanning diverse NLP tasks, as well as a real-world undocumented HuggingFace model. Results show that label forensics consistently reveals label semantics, detects mismatches between declared and learned behaviors, and supports responsible AI auditing.
\end{itemize}

\section{Related Works}


While several attack techniques in NLP privacy and security inspire components of our pipeline~\cite{llmattacksurvey,chowdhury2024breaking,llmattacksurvey2}, our goal fundamentally differs from theirs. Membership inference attacks~\cite{shokri2017membership,choquette2021label,duan2024membership} and embedding inversion~\cite{chen2024embedding, li2023sentenceembedding} demonstrate that sensitive data can leak through language model outputs. Other efforts develop targeted data-reconstruction attacks aimed at recovering representative training inputs from text classifiers~\cite{elmahdy2023classifierconstructing}. In addition to these reconstruction-oriented methods, training data extraction attacks, originally developed for generative models, have demonstrated that it is possible to scale up the recovery of training content from deployed models~\cite{icmlCarlini, nasr2025scalable, dentan2024reconstructing}. While our method includes data generation for semantic analysis, our objective is not to reconstruct specific training samples for privacy leakage. Instead, we aim to generate sparse and interpretable samples that support the recovery of precise and general label concepts, rather than reproducing individual training instances.


The interpretability or explainability of language models also aligns with our task~\cite{interpret, wang2023largeinterpret, yang-etal-2024-enhancinginterpret,ehsan2024humaninterpret}. Recent studies have explored internal mechanisms within language models using sparse autoencoders to extract thousands of human-interpretable features~\cite{belrose2023sparse,neo2024towards}, or applying causal interventions to identify robust causal pathways for prediction~\cite{kiciman2023causal,huang2025internal}. Beyond internal probing, black-box interpretability methods have also emerged~\cite{singh2023explaining,oikarinen2023labelfree,wen2025language,chen2025automated,rauba2025auditing}. Existing approaches usually explain a model behavior by analyzing its inputs and outputs, highlighting important tokens, or approximating the local decision boundary around individual examples. While these methods improve transparency, they mostly clarify why a single prediction was made rather than revealing the broader semantic patterns or label-level structures that the model has learned.


Auditing language models in opaque settings has focused on probing functional behaviors such as refusal patterns~\cite{arditi2024refusal, auditllm2024}, jailbreak susceptibility~\cite{jailbreak}, or knowledge tracing~\cite{marks2025auditing}. These approaches evaluate how the model behaves on harmful or safety-critical tasks. However, they do not examine what each output label actually represents. In contrast, label forensics focuses on the semantic content of individual labels, aiming to recover the underlying concept a classifier associates with each label rather than assessing its high-level behavioral properties.

While the task of label forensics is underexplored in NLP, conceptually related efforts in computer vision, such as label inference attacks and domain discovery~\cite{fucong,zhao2025personalized,domain}. Yet natural language inputs differ significantly from images: they are discrete, compositional, and semantically entangled, which makes direct transfer of such techniques non-trivial.

\section{Problem Formulation}

\subsection{Capabilities of the Investigator}

Consider an investigator with black-box and hard-label only access to a text classification model $\mathcal{M}: \mathcal{X} \rightarrow \mathcal{Y}$, where $\mathcal{X}$ denotes the space of natural language sentences and $\mathcal{Y}$ is a finite set of class labels. The model $\mathcal{M}$ is accessible only via queries, as investigators have no access to model internal parameters, architectural details, or gradients. Moreover, $\mathcal{M}$ returns hard labels only, which do not expose confidence scores, probability distributions, or logits. All queries must be issued in the form of natural language sentences.

The investigator is equipped with several auxiliary resources. An open-source large language model $\mathcal{G}$ (e.g., Llama or the Qwen series) provides a flexible mechanism for producing diverse natural language sentences. An external corpus offers additional knowledge, including a lexical hierarchy $\mathcal{W}$ derived from \texttt{WordNet} and a collection of harmful or safety-related sentences, which provides coverage of linguistic regions that are difficult to obtain through $\mathcal{G}$ alone. Finally, the investigator has a sampler $\mathcal{S}$ implemented as a prompt-free encoder–decoder pair $(\texttt{Enc}, \texttt{Dec})$, which maps a sentence to a continuous semantic embedding and stochastically decodes embeddings back into natural language. 

\subsection{Goal of Investigation}

For each label $y \in \mathcal{Y}$, the investigator aims to recover the semantic concept that the classifier $\mathcal{M}$ implicitly associates with that label. We formalize this concept as a semantic distribution $\mathcal{D}_{y}$ over natural language sentences or their embeddings, representing the region of meaning for which the classifier consistently predicts $y$.

We formalize the task as approximating a semantic distribution $\mathbb{D}_y= \{(e_1, \alpha_1), \ldots, (e_m, \alpha_m)\},$ where each embedding $e_i$ corresponds to a representative sentence associated with label $y$ and each $\alpha_i$ characterizes the size of the semantic neighborhood around $e_i$ within which $\mathcal{M}$ continues to predict the same label. Intuitively, the distribution $\mathbb{D}_y$ defines a region of natural language expressions that are both consistently classified as label $y$ and semantically representative of the label. 

A well-formed distribution $\mathbb{D}_y$ should satisfy two key properties:
\begin{itemize}
    \item \textbf{Precise:} Sentences sampled from $\mathbb{D}_y$ must be reliably classified by $\mathcal{M}$ as label $y$:
    \begin{equation}
        \forall x \in \mathrm{supp} \;\mathbb{D}_y,\quad \mathcal{M}(x) = y.
    \end{equation}
    \item \textbf{General:} The embeddings in $\mathbb{D}_y$ should form a broad and coherent semantic region. Formally, we define $\mathbb{E}_{e_i,\,e_j \in \mathbb{D}_y}\big[ d(e_i, e_j) \big]\quad ,$ where $d$
    for all $y' \neq y$ where $d$ denotes a distance measure in the embedding space, and a larger value indicates a wider semantic spread.
\end{itemize}

\subsection{Objective Function}

For each label $y$, we operationalize these two properties by optimizing the following objective:

\begin{equation}
\max_{\mathbb{D}_y}\; \Pr_{x \in \mathrm{supp} \; \mathbb{D}_y }[\mathcal{M}(x)=y] \;+\; \lambda\, \mathbb{E}_{e_i,\,e_j \in \mathbb{D}_y}\big[d(e_i, e_j)\big].
\end{equation}

The two terms jointly ensure that $\mathbb{D}_y$ remains both decision-aligned and semantically expansive, yielding a distribution that reflects the full concept associated with label $y$.

\section{Proposed Framework}

\begin{figure}[t]
\centering
\includegraphics[width=0.9\columnwidth]{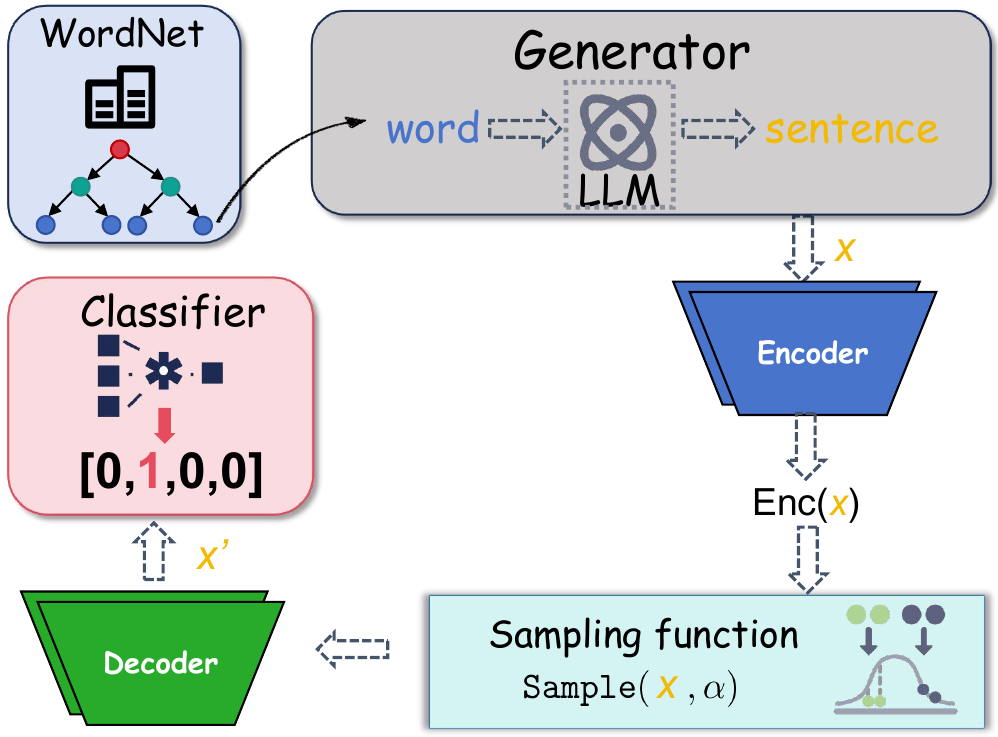} 
\caption{The semantic distribution construction pipeline. Sentences are generated from \texttt{WordNet} using an LLM. An encoder–decoder architecture serves as the semantic neighborhood sampler. Each seed sentence is paired with a sampling radius $\alpha$ to define a controllable semantic distribution.}
\label{fig2}
\end{figure}

The label forensics pipeline consists of three stages: semantic distribution initialization, prompt-free sampler construction with iterative optimization, and semantic interpretation. Figure~\ref{fig2} provides a simplified illustration of the process for constructing a label-specific semantic distribution. We begin by generating a candidate prototype sentence set $\mathcal{A}_y$ for each label using an open-source LLM conditioned on lexical anchors extracted from corpora such as WordNet. This forms the initial semantic support for label $y$. To approximate the distribution $\mathbb{D}_y$, we design a sampling mechanism based on $\mathcal{A}_y$. The sampling mechanism uses a prompt-free encoder–decoder sampler $(\texttt{Enc}, \texttt{Dec})$ where the encoder $\texttt{Enc}$ maps $x \in \mathcal{A}_y$ to a continuous embedding $\mathbf{e}$. The decoder $\texttt{Dec}$ reconstructs a rephrased variant $x^\prime = \texttt{Dec}(\mathbf{e})$ from the latent embedding.


\subsection{Semantic Distribution Initialization}

Recovering the semantic distribution $\mathbb{D}_y$ for a label $y$ involves identifying a set of natural language sentences that are both consistently classified as $y$ and semantically representative of the class. Unlike prior work on data reconstruction attacks, these sentences are not intended to replicate training examples, but rather to capture the underlying semantics of each label. The initialization process begins by collecting a diverse set of prototype sentences and constructing a latent-space sampling mechanism. In the following, we elaborate on each component in detail, describing how the sampler is optimized and how the resulting samples are used to recover the semantic distribution.

\subsubsection{Heuristic Hierarchical Traversal over \texttt{WordNet}.}

To obtain an initial pool of semantically prototype sentences, lexical words are first extracted from \texttt{WordNet}, a structured semantic graph composed of synsets and hypernymy relations. Beginning from root-level abstract words, a breadth-first traversal explores increasingly specific word-level nodes. For each anchor word $w$, a sentence $x \sim \mathcal{G}(\pi(w))$ is generated using a language model $\mathcal{G}$ under a fixed prompt template $\pi(\cdot)$. The sentence is submitted to the black-box classifier $\mathcal{M}$, and retained as part of the candidate prototype sentence set $\mathcal{A}_y$ if $\mathcal{M}(x) = y$. This procedure yields a class-conditional set of diverse anchor sentences:
\begin{equation}
    \mathcal{A}_y = \big\{\, x \sim \mathcal{G}(\pi(w)) \;\big|\; w \in \mathcal{W},\; \mathcal{M}(x) = y \,\big\}, 
\end{equation}

While the \texttt{WordNet}-based initialization is effective for most classes, certain labels such as \texttt{jailbreak} remain underrepresented due to class imbalance. To mitigate this issue, a three-path expansion strategy is adopted:
\begin{itemize}
    \item \textbf{Word-level Expansion.} For each word $w$ associated with the candidate prototype $x \in \mathcal{A}_y$, $w^{\prime}$ are retrieved using \texttt{WordNet} relations such as \texttt{similar\_to} and \texttt{also\_see}. Each $w^{\prime}$ is passed to the generator $\mathcal{G}$ to produce a new sentence $x^{\prime} = \mathcal{G}(\pi(w^{\prime}))$. If $\mathcal{M}(x^{\prime}) = y$, the sample $x^{\prime}$ is added to $\mathcal{A}_y$.
    \item \textbf{Sentence-level Expansion.}  Each $x \in \mathcal{A}_y$ is paraphrased into a set ${x^{\prime}_i}$ using $\mathcal{G}$ to explore semantically equivalent sentences. Those satisfying $\mathcal{M}(x^{\prime}_i) = y$ are retained.
    \item \textbf{External Corpus Expansion.}  A curated corpus of harmfuly-related sentences is additionally included to cover toxic regions as the LLM $\mathcal{G}$ is hard to generate toxic or harmful content. Sentences classified as $y$ are added to $\mathcal{A}_y$.
\end{itemize}

\subsubsection{Prompt-free Sampler Construction.}

To explore the semantic distribution $\mathbb{D}_y$, we construct a latent-space neighborhood sampler based on a prefix-tuned encoder–decoder architecture.\footnote{Decoder-only models such as GPT lack encoder cross-attention and cannot condition directly on external embeddings.} The sampling mechanism uses a prompt-free sampler $(\texttt{Enc}, \texttt{Dec})$ where the encoder $\texttt{Enc}$ maps $x \in \mathcal{A}_y$ to a continuous embedding $\mathbf{e} = \texttt{Enc}(x) \in \mathbb{R}^{l \times d}$, where $l$ is the sequence length of $x$ and $d$ is the embedding dimension. The decoder $\texttt{Dec}$ reconstructs a rephrased variant $x^\prime = \texttt{Dec}(\mathbf{e})$ from the latent embedding. To draw samples $x^\prime \sim \mathbb{D}_y$, we first sample a candidate prototype sentence uniformly from $\mathcal{A}_y$ then apply Gaussian noise in its embedding space so that the output sentence inherits the semantic concept but differs in concrete expression. The stochastic sampling function is defined as:
\begin{equation}
\label{eq:sample}
\begin{aligned}
    x^\prime &= \texttt{Dec}(\texttt{Enc}(x) + \delta), \\
    x &\sim \mathcal{U}(\mathcal{A}_y),\quad \delta \sim \mathcal{N}(0, \alpha^2 I).
\end{aligned}
\end{equation}

Here, $\mathcal{U}(\cdot)$ means uniformly sampling and $\alpha$ controls the scale of perturbation in embedding space.

A key requirement is prompt-free generation: the decoder must generate solely from embeddings of prototype sentences, without reliance on handcrafted prompts. However, the standard T5 models rely on textual prompts (e.g., \texttt{\underline{paraphrase:} I am happy}) to control task-oriented generation. To remove this dependency, we fine-tune T5 using prefix tuning, optimizing only a small set of continuous virtual prefix vectors while keeping the backbone frozen. Specifically, we build on the publicly available model\footnote{\url{https://huggingface.co/humarin/chatgpt_paraphraser_on_T5_base}} and fine-tune it on the ChatGPT-generated dataset\footnote{\url{https://huggingface.co/datasets/humarin/chatgpt-paraphrases}} to mimic prompt-based behavior in a prompt-free setting. 

The training objective aligns the prompt-free and prompt-based models using the following distillation loss:
\begin{equation}
    \mathcal{L}_{\text{align}} = \mathbb{E}_{x \sim \mathcal{D}} \left[ \mathrm{CE}\left( \mathcal{T}(\cdot \mid \pi(x)),\; \mathcal{T}_{\text{prefix}}(\cdot \mid x) \right) \right],
\end{equation}
where $\mathcal{T}$ is the original T5 model with prompt $\pi(x)$, and $\mathcal{T}_{\text{prefix}}$ is the prefix-tuned variant without prompts. The learned prefix enables fluent, semantically consistent decoding from latent inputs, and serves as the backbone for controlled perturbation-based sampling.

\subsubsection{Sampling Radius Estimation.}

The \textit{sampling radius} $\alpha$ characterizes the maximum semantic perturbation under which the model consistently classifies a prototype sentence $s$ with probability at least $\eta$. Formally, it is defined as:
\begin{align}
\alpha(s, \eta) = \sup \Big\{ \alpha \,\Big|\, 
\mathbb{P}_{\delta \sim \mathcal{N}(0, \alpha^2 I)} 
\big[&\mathcal{M}(\texttt{Dec}(\texttt{Enc}(s) + \delta)) \notag \\
&= \mathcal{M}(s)
\big] \geq \eta \Big\}.
\end{align}

Here, $\eta \in (0, 1)$ denotes a confidence threshold (e.g., $\eta = 0.7$), ensuring that sampled variants retain the original label with high probability. A large $\alpha(s, \eta)$ indicates that the prototype lies at the center of a semantically stable region: the classifier’s output remains consistent even under strong latent perturbations. 

The sampling radius $\alpha(s, \eta)$ is estimated via Monte Carlo sampling with binary search: for each candidate $\alpha$, we draw $m$ perturbations from $\mathcal{N}(0, \alpha^2 I)$, decode each perturbed embedding, and query the classifier. If the label match rate exceeds $\eta$, the search radius is expanded; otherwise, it is reduced. This continues until convergence, yielding a robustness estimate $\alpha(s, \eta)$ per sentence (see Listing~\ref{lst:alpha_estimation}).

\begin{listing}[tb]%
\caption{Estimating sampling radius via binary search}%
\label{lst:alpha_estimation}%
\begin{lstlisting}[language=Python]
def estimate_alpha(s, eta, m):
    Z = Enc(s); y = M(Dec(Z))
    low, high = alpha_min, alpha_max
    eps = 1e-3
    while high - low > eps:
        alpha = (low + high) / 2
        count = 0
        for _ in range(m):
            delta = Gaussian(std=alpha)
            y_hat = M(Dec(Z + delta))
            count += (y_hat == y)
        if count / m >= eta:
            low = alpha
        else:
            high = alpha
    return low
\end{lstlisting}
\end{listing}

\subsection{Prototype Scoring Objective}

To recover a precise and general distribution $\mathbb{D}_y$ for each class $y$, we introduce a scoring objective to optimize prototype sentences $s \in \mathcal{A}_y$. In addition to the sampling radius $\alpha(s, \eta)$, which reflects classification consistency under perturbations, we incorporate two embedding-based criteria that further identify semantically representative samples: \textit{consistency} and \textit{separability}.

Specifically, let $\mathbf{z} \in \mathbb{R}^d$ denote the sentence embedding for $s$, computed via mean pooling over the final encoder hidden states $\mathbf{z} = \frac{1}{l} \sum_{i=1}^l \mathbf{h}_i$, where $\{\mathbf{h}_i\}_{i=1}^{L}$ are the token embeddings from the encoder $\texttt{Enc}$. Given a predicted label $y$, the centroid of class $\mathbf{c}_y$ is computed as the mean of normalized sentence embeddings in $\mathcal{A}_y$. We define:
\begin{itemize}
    \item \textbf{Consistency}: Measures how well a sample aligns with the core semantics of its predicted class, defined as the cosine similarity between $\mathbf{z}$ and its class centroid $\mathbf{c}_y$;
    \item \textbf{Separability}: Penalizes proximity defined as the inverse of the maximum cosine similarity between $\mathbf{z}$ and any other class centroid $\mathbf{c}_{\tilde{y}}$, where $\tilde{y} \in \mathcal{Y}\setminus\{y\}$.
\end{itemize}

These two metrics are designed to ensure that selected samples are semantically faithful to their class while remaining distinct from others. The overall scoring function is:  
\begin{equation}  
    \mathcal{R}(s) = \alpha(s,\eta) + \lambda \cdot \cos(\mathbf{z}, \mathbf{c}_y) + \gamma \cdot \left(1 - \max_{\tilde{y} \in \mathcal{Y}\setminus\{y\}} \cos(\mathbf{z}, \mathbf{c}_{\tilde{y}})\right),  
\end{equation}  
where $\lambda, \gamma \ge 0$ control the balance between consistency and separability. 

The optimization objective is to select a subset of $K$ prototype sentences $\mathcal{S}_y \subset \mathcal{A}_y$ that maximizes the total representativeness:

\begin{equation}
\label{eq:subset_selection}
\mathcal{S}_y^\star
= \operatorname*{arg\,max}_{{|\mathcal{S}_y|=K}}
\;\; \sum_{s \in \mathcal{S}_y} \mathcal{R}(s).
\end{equation}

The selected subset $\mathcal{S}_y^\star$ provides a compact and interpretable summary of the semantic space associated with label $y$, and serves as the support for the induced distribution $\mathbb{D}_y$, where each prototype $s \in \mathcal{S}_y^\star$ is associated with a sampling radius $\alpha(s, \eta)$.



\subsection{Interpreting the Semantic Distribution}

After constructing the semantic distribution for each label $y \in \mathcal{Y}$, the next objective is to distill high-level, interpretable descriptions that summarize the label’s semantics in a human-understandable form. Specifically, the goal is to extract abstract words or phrases that generalize across diverse prototype sentences. For example, in an emotion classifier, rather than listing expressions like \texttt{I can’t stop smiling} or \texttt{What a wonderful day}, we aim to induce canonical concepts such as \texttt{joy} that capture the underlying labeling behavior of the model.

As $\mathbb{D}_y$ encodes the text classifier’s behavior across semantically coherent inputs, we can directly prompt a generative language model $\mathcal{G}$ to attribute an interpretable label description. Specifically, the model is asked to generate a set of candidate label descriptions $\{d_1, \ldots, d_k\}$ that summarize the shared semantics of $\mathcal{S}_y^\star$. Each $d_i$ is expected to be a short phrase that captures the essence of class $y$.

To evaluate the alignment between each candidate $d_i$ and the semantic behavior of the classifier, we employ a pretrained natural language inference (NLI) model as a scoring function. For each prototype sentence $s \in \mathcal{S}_y^\star$, we compute the entailment score $f_{\text{NLI}}(s, d_i)$. The overall quality of label description is quantified by its entailment hit rate:

\begin{equation}
\label{eq:calculate_entail_score}
    \mathcal{H}(d_i) = \frac{1}{|\mathcal{S}_y^\star|} \sum_{s \in \mathcal{S}_y^\star} \mathbb{I} \left[ f_{\text{NLI}}(s, d_i) \geq \tau \right],
\end{equation}

where $\tau$ is a confidence threshold (e.g., $\tau = 0.6$), and $\mathbb{I}[\cdot]$ is the indicator function. This score measures how often the prototype sentences entail the candidate label description according to the NLI model. Label descriptions with higher $\mathcal{H}(d_i)$ are considered more faithful and class-representative. This mechanism provides an interpretable summarization of each label’s semantic footprint in terms of human-readable descriptions, bridging the gap between black-box classifier predictions and semantic transparency.

\section{Experimental Results}

\subsection{Implementation Details}

We evaluate our method on five publicly available black-box text classifiers, each addressing a distinct NLP task:

\begin{itemize}
    \item \textbf{Emotion Classification}($\mathcal{M}_{\text{emo}}$): a DistilRoBERTa-based model trained to classify text into one of seven emotion categories.
    \item \textbf{Sentiment Analysis} ($\mathcal{M}_{\text{sent}}$): a multilingual BERT model that assigns a sentiment rating from 1 (most negative) to 5 (most positive). Since the model lacks explicit label descriptions, it presents a natural scenario for unsupervised label forensics.
    \item \textbf{Spam Detection} ($\mathcal{M}_{\text{spam}}$): a binary classifier trained on corpora of emails and SMS messages to distinguish spam from legitimate content.
    \item \textbf{Jailbreak Detection} ($\mathcal{M}_{\text{jail}}$): a DistilBERT model designed to identify adversarial jailbreak prompts.
    \item \textbf{Toxicity Detection} ($\mathcal{M}_{\text{tox}}$): a RoBERTa-based classifier for binary toxicity classification.
\end{itemize}

All models in our study operate under a strict black-box setting, exposing only hard-label predictions without access to confidence scores or logits. To construct prototype sentence candidates, we traverse the lexical graph of \texttt{WordNet 3.1} using hierarchical heuristics. For probing and concept induction, we utilize the open-source large language models \texttt{Qwen-Chat} and \texttt{Llama-3.1-8B-Instruct} as the generator $\mathcal{G}$, configured with nucleus sampling ($p = 0.9$) and temperature $1.0$ to balance output diversity and semantic fidelity.

For training the prompt-free neighborhood sampler, we adopt prefix tuning by appending $20$ learnable virtual tokens to both the encoder and decoder. The tuning follows a teacher–student distillation setup and is performed for $2$ epochs with a batch size of $64$ and a learning rate of $10^{-4}$. For each label $y$, to assess the semantic alignment of induced concepts, we use an entailment model $f_{\text{NLI}}$, with the entailment confidence threshold $\tau$ set to $0.6$.


All experiments are conducted on a workstation equipped with 4 NVIDIA A40 GPUs (46 GB VRAM each).

\subsection{Evaluation Setup}

To assess the structural quality of the induced semantic distribution, we evaluate the geometry of prototype sentence embeddings using two metrics: intra-class distance and inter-class distance. All embeddings are derived from the selected prototype subsets $\mathcal{S}_y^\star$ for each class $y \in \mathcal{Y}$. Specifically, we compute sentence embeddings $\mathbf{z} \in \mathbb{R}^D$ using pretrained encoders such as Sentence-BERT and SimCSE to ensure model-agnosticity. Let $\mathcal{Z}_y = \{\mathbf{z}_1, \ldots, \mathbf{z}_{n_y}\}$ denote the embedding set obtained from $\mathcal{S}_y^\star$. The centroid of each class is defined as:
\begin{equation}
\mathbf{c}_y = \frac{1}{|\mathcal{Z}_y|} \sum_{\mathbf{z} \in \mathcal{Z}_y} \mathbf{z}.
\end{equation}

The intra-class distance measures the average distance between samples and their class centroid:
\begin{equation}
\label{eq:d_intra_code} 
d_{\text{intra}} = \frac{1}{|\mathcal{Y}|} \sum_{y \in \mathcal{Y}} \left( \frac{1}{|\mathcal{Z}_y|} \sum_{\mathbf{z} \in \mathcal{Z}_y} \left(1 - \cos(\mathbf{z}, \mathbf{c}_y) \right) \right).
\end{equation}

The inter-class distance computes the average distance between all pairs of class centroids:
\begin{equation}
\label{eq:d_inter_code} 
d_{\text{inter}} = \frac{1}{\binom{|\mathcal{Y}|}{2}} \sum_{y_i < y_j} \left(1 - \cos(\mathbf{c}_{y_i}, \mathbf{c}_{y_j}) \right).
\end{equation}

Low $d_{\text{intra}}$ indicates that prototype samples are semantically coherent within each class, while high $d_{\text{inter}}$ suggests that the induced classes are geometrically well-separated in embedding space.

To evaluate the quality of the generated semantic distributions, we adopt BERTScore and Self-BLEU. BERTScore assesses semantic similarity between generated samples and their corresponding prototypes, capturing meaning beyond surface-level tokens. In contrast, Self-BLEU measures intra-set diversity; lower values indicate greater variation among samples, helping to identify redundancy or mode collapse.


\begin{table}[t]
\centering
\small
\begin{tabular}{lccccc}
\toprule
\textbf{Label} &
\textbf{Emotion} &
\textbf{Sentiment} &
\textbf{Spam} &
\textbf{Jailbreak} &
\textbf{Toxicity} \\
\midrule
0 & 86.0 & 93.0 & 99.0 & 100.0 & 100.0 \\
1 & 93.0 & 86.0 & 93.0 & 84.0  & 91.0  \\
2 & 95.0 & 82.0 & --   & --    & --    \\
3 & 93.0 & 89.0 & --   & --    & --    \\
4 & 88.0 & 83.0 & --   & --    & --    \\
5 & 91.0 & --   & --   & --    & --    \\
6 & 92.0 & --   & --   & --    & --    \\
\midrule
Avg. & 91.1 & 86.6 & 96.0 & 92.0 & 95.5 \\
\bottomrule
\end{tabular}
\caption{Accuracy (\%) of classifier predictions on 100 sampled sentences per label generated from the reconstructed embedding space.}
\label{tab:numerical_results_reordered}
\end{table}

\subsection{Quantitative Results}


\subsubsection{Label Consistency in Reconstructed Embedding Space.}
We first assess the fidelity of the reconstructed embedding space.
For each label-specific embedding space, we uniformly sample 100 embeddings, decode them into sentences, and evaluate whether the classifier assigns the original label. A higher accuracy indicates that the learned embedding space faithfully captures the classifier’s decision manifold, preserving both local smoothness and label-level semantic coherence. As shown in Table~\ref{tab:numerical_results_reordered}, across all five classifiers, the recovered embedding regions exhibit strong label consistency. Binary models such as toxicity (95.5\%) and spam (96.0\%) show almost perfectly coherent embedding neighborhoods, confirming that their decision boundaries are sharp and well-separated. The jailbreak detector yields a mean consistency of 92.0\%, though its harmful-content region (Label 1: 84\%) is less stable, reflecting the inherently irregular boundary around safety-critical content. Multi-class models show similarly robust behavior. The emotion classifier achieves 91.1\% average consistency across seven labels, and the sentiment classifier maintains 86.6\% despite its finer-grained five-way label space. These results demonstrate that the embedding reconstruction process successfully recovers classifier-aligned semantic regions that remain internally coherent under sampling-based perturbations.

\subsubsection{Semantic Alignment with Ground-Truth Labels.}

\begin{table}[t]
\centering
\begin{tabular}{l|l|l}
\toprule
\textbf{Task} & \textbf{Label} & \textbf{Inferred Concepts} \\
\midrule
\multirow{7}{*}{Emotion} 
    & Anger      &  anger, hatred\\
    & Disgust    &  disgust, discomfort\\
    & Fear       &  fear, danger\\
    & Joy        &  joy, cheerfulness\\
    & Neutral    &  standards, balance\\
    & Sadness    &  sadness, heartache\\
    & Surprise   &  astonishment, surprise \\
\midrule
\multirow{5}{*}{Sentiment}
    & Very negative   &  theft, fraud\\
    & Negative        &  fatigue, weakened\\
    & Neutral         &  conventional, tradition\\
    & Positive        &  effectiveness, effort\\
    & Very positive   &  successful, happiness\\
\midrule
\multirow{2}{*}{Spam}
    & Ham       &  brightness, cleanliness\\
    & Spam      &  irrationality, trustlessness\\
\midrule
\multirow{2}{*}{Jailbreak}
    & Clean Prompt    &   brightness, relaxation  \\
    & Jailbreak       &   fulfillment, prerequisites\\
\midrule
\multirow{2}{*}{Toxicity}
    & Non-Toxic       &  responsibility, health\\
    & Toxic           &  irrationality, harm\\
\bottomrule
\end{tabular}
\caption{Representative label descriptions inferred for each label across five black-box classifiers.}
\label{tab:concept_fidelity}
\end{table}

To bridge the gap between semantic distribution and human interpretability, we distill the recovered distributions into concise, natural language label descriptions. Table~\ref{tab:concept_fidelity} reports the inferred descriptions across five black-box classifiers spanning diverse NLP tasks.

The quality of inferred label descriptions varies across tasks. For emotion classification, the generated descriptions exhibit nearly one-to-one correspondence with the true emotion categories (e.g., \texttt{joy, anger, fear}), indicating that the semantic distributions accurately capture label semantics. In sentiment analysis, despite the model providing only numeric labels (1–5) without textual descriptions, the induced descriptions align closely with sentiment polarity. For tasks with less well-defined or inherently subjective label semantics, such as spam detection, jailbreak detection, and toxicity classification, the recovered descriptions still exhibit coherent and interpretable behavioral patterns. For example, the \texttt{spam} label is associated with terms such as \texttt{irrationality} and \texttt{trustlessness}, while the \texttt{toxic} label yields descriptors like \texttt{harm}, which align with human intuition. Jailbreak detection, being the least well-defined, produces less precise yet still informative descriptions (e.g., \texttt{fulfillment, prerequisites}), which capture the general intent of adversarial prompts.

\begin{table}[t]
\centering
\begin{tabular}{llccc}
\toprule
\textbf{Task} & \textbf{Model} & $d_{\text{intra}}$ ↓ & $d_{\text{inter}}$ ↑ & $r$ ↑ \\
\midrule
\multirow{2}{*}{Emotion}    
    & SBERT   & 0.7766 & 0.9193 & 1.18 \\
    & SimCSE  & 0.4766 & 0.5959 & 1.25 \\
\midrule
\multirow{2}{*}{Sentiment} 
    & SBERT   & 0.8802 & 0.9376 & 1.07 \\
    & SimCSE  & 0.5330 & 0.5954 & 1.12 \\
\midrule
\multirow{2}{*}{Spam}      
    & SBERT   & 0.5210 & 0.9977 & 1.92 \\
    & SimCSE  & 0.3455 & 0.6152 & 1.78 \\
\midrule
\multirow{2}{*}{Jailbreak} 
    & SBERT   & 0.6128 & 1.0250 & 1.67 \\
    & SimCSE  & 0.4110 & 0.6503 & 1.58 \\
\midrule
\multirow{2}{*}{Toxicity}  
    & SBERT   & 0.7632 & 0.9470 & 1.24 \\
    & SimCSE  & 0.4941 & 0.6258 & 1.27 \\
\bottomrule
\end{tabular}
\caption{Intra-class and inter-class distances of sentence embeddings across models. Lower $d_{\text{intra}}$ and higher $d_{\text{inter}}$ indicate better geometric structure. The ratio $r = d_{\text{inter}} / d_{\text{intra}}$ quantifies overall class separability.}
\label{tab:embedding-geometry}
\end{table}

\begin{table}[t]
\centering
\begin{tabular}{lcccc}
\toprule
\textbf{Task} & \textbf{BERTScore} ↑ & \textbf{Consistency}  ↑ & \textbf{BLEU} ↓ \\
\midrule
Emotion      & 0.94 & 100.0\% & 0.22  \\
Sentiment    & 0.95 & 100.0\% & 0.18  \\
Spam         & 0.93 & 100.0\% & 0.23  \\
Jailbreak    & 0.93 & 100.0\% & 0.28  \\
Toxicity     & 0.94 & 100.0\% & 0.16  \\
\bottomrule
\end{tabular}
\caption{Evaluation of neighborhood sampler over prototype sentences $\mathcal{S}_y^\star$ across tasks. BERTScore captures semantic fidelity, Consistency measures label preservation, and BLEU reflects lexical diversity.}
\label{tab:t5-generation-eval}
\end{table}


\subsubsection{Geometric Analysis of Prototype Sentences.}

\begin{figure*}[t] 
\centering
\subfigure[Emotion $\mathcal{M}_{\text{emo}}$]{
\includegraphics[width=3.3cm]{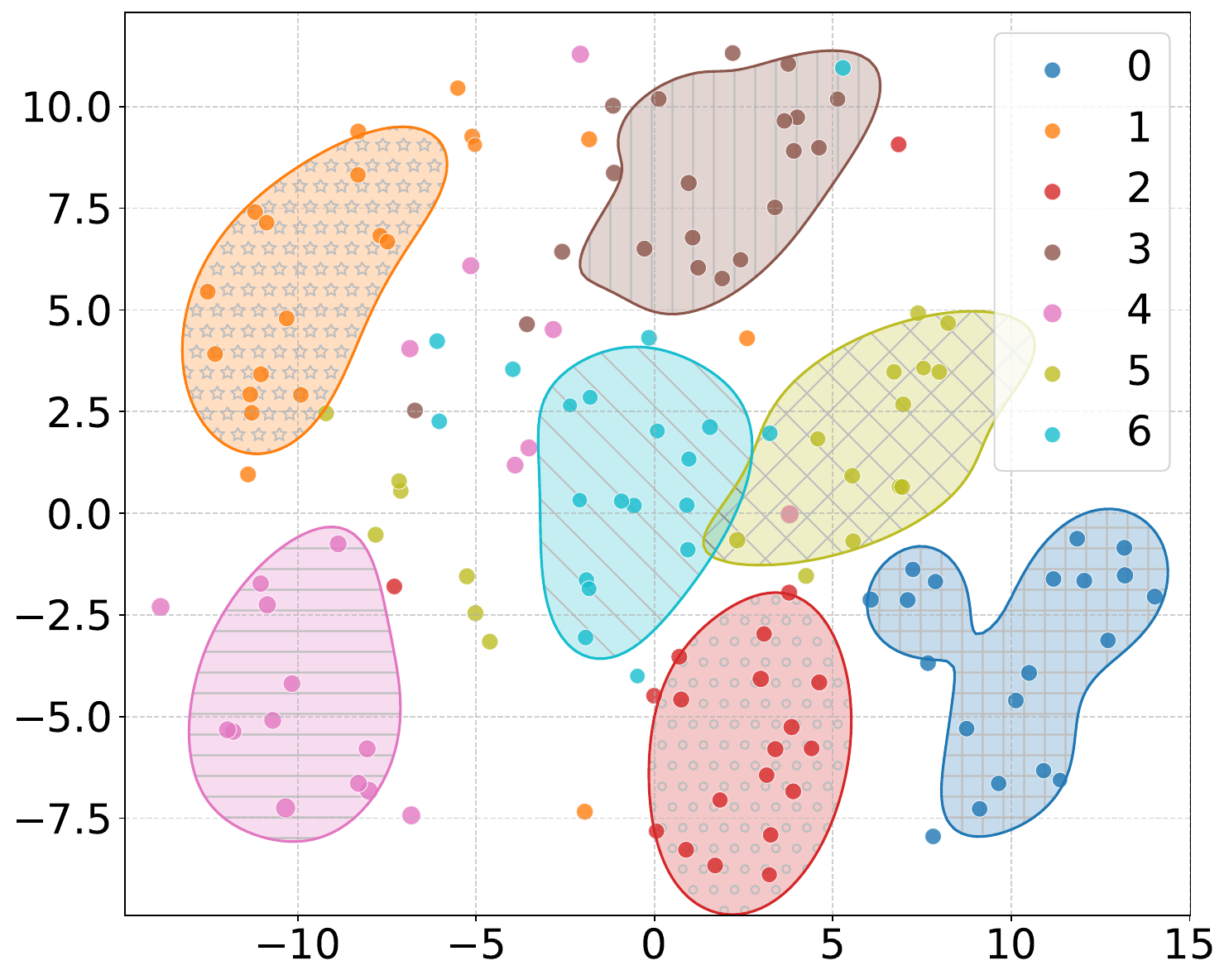}
} 
\subfigure[Sentiment $\mathcal{M}_{\text{sent}}$]{
\includegraphics[width=3.3cm]{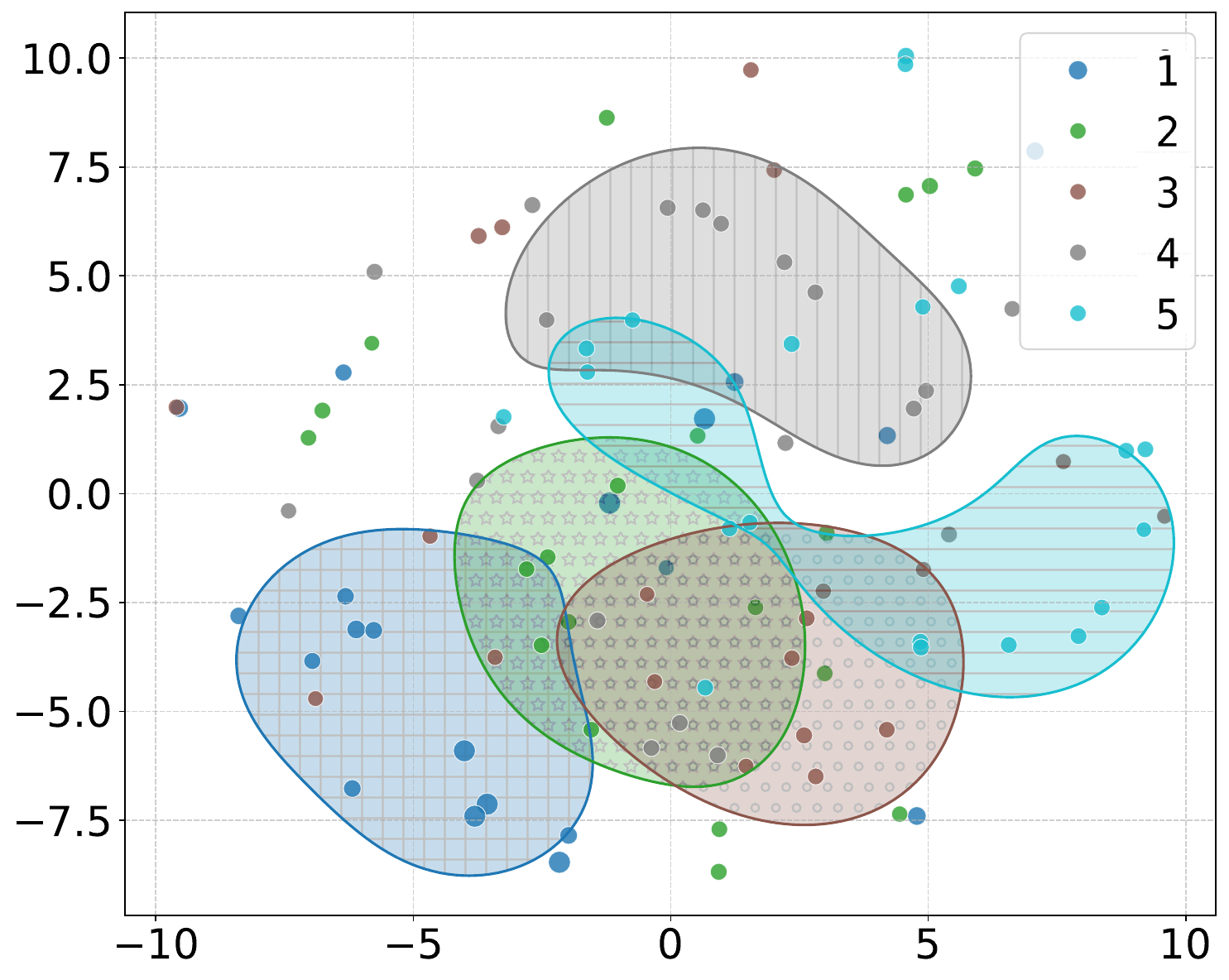}
}
\subfigure[Spam $\mathcal{M}_{\text{spam}}$]{
\includegraphics[width=3.3cm]{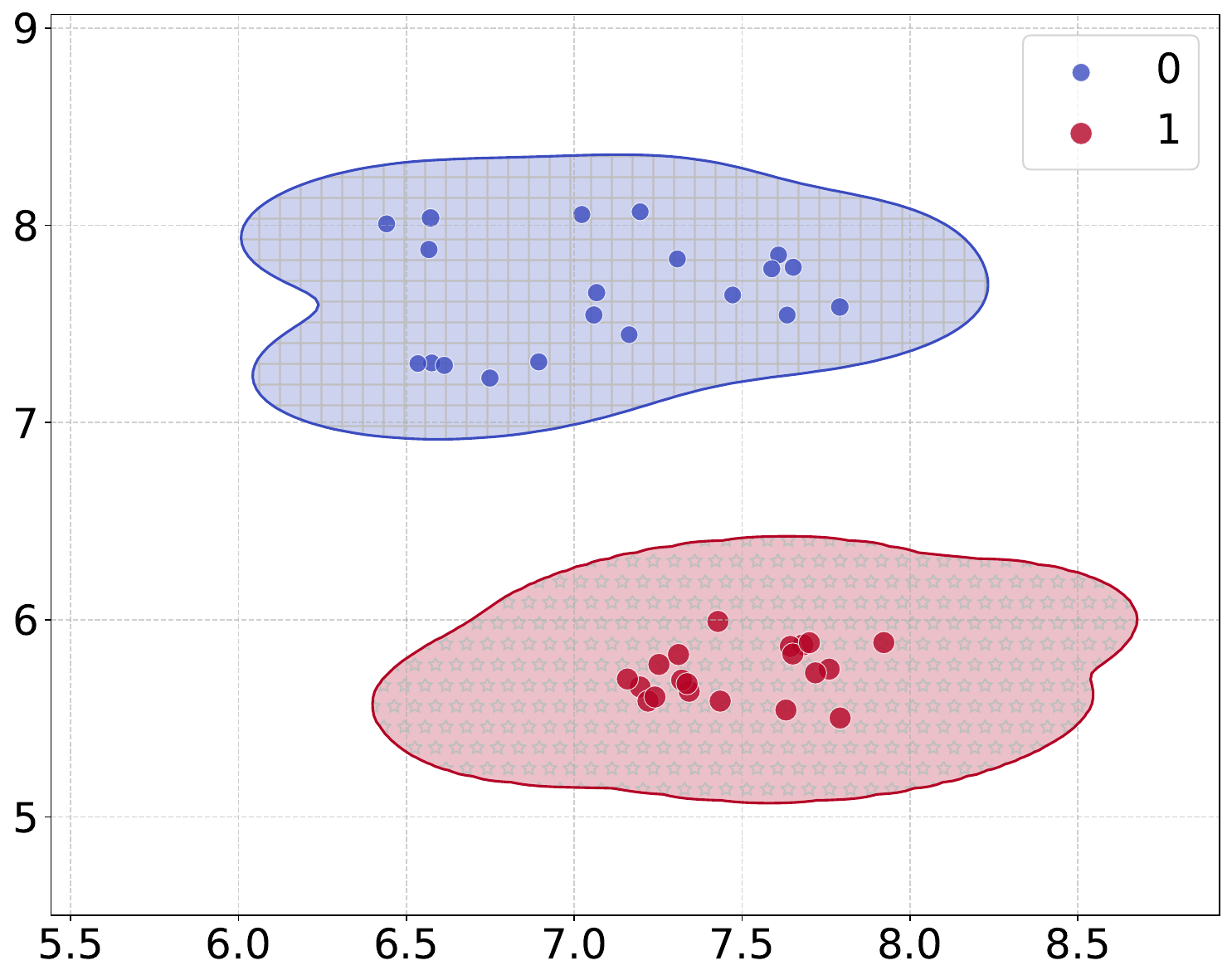}
}
\subfigure[Jailbreak $\mathcal{M}_{\text{emo}}$]{
\includegraphics[width=3.3cm]{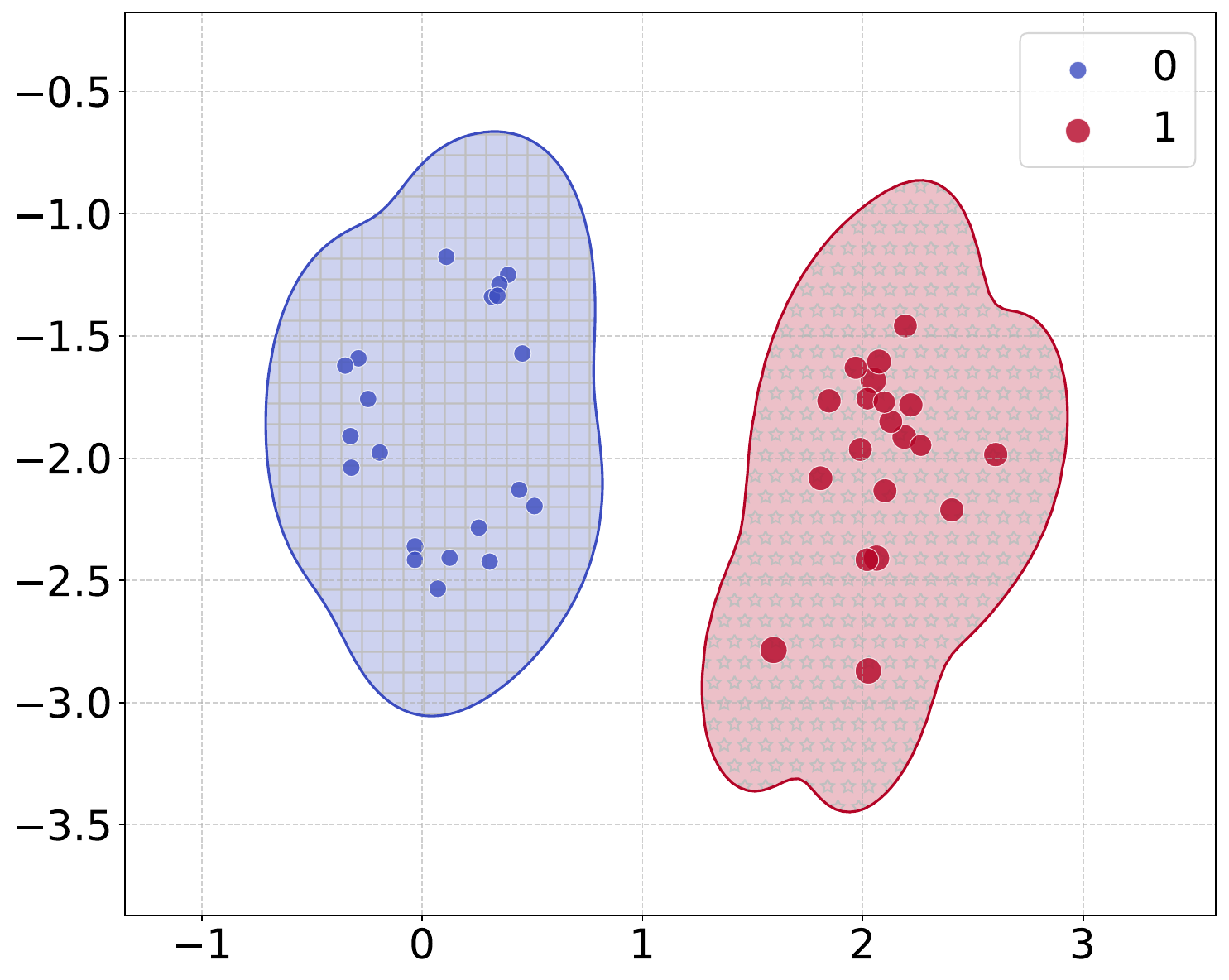}
}
\subfigure[Toxicity $\mathcal{M}_{\text{emo}}$]{
\includegraphics[width=3.3cm]{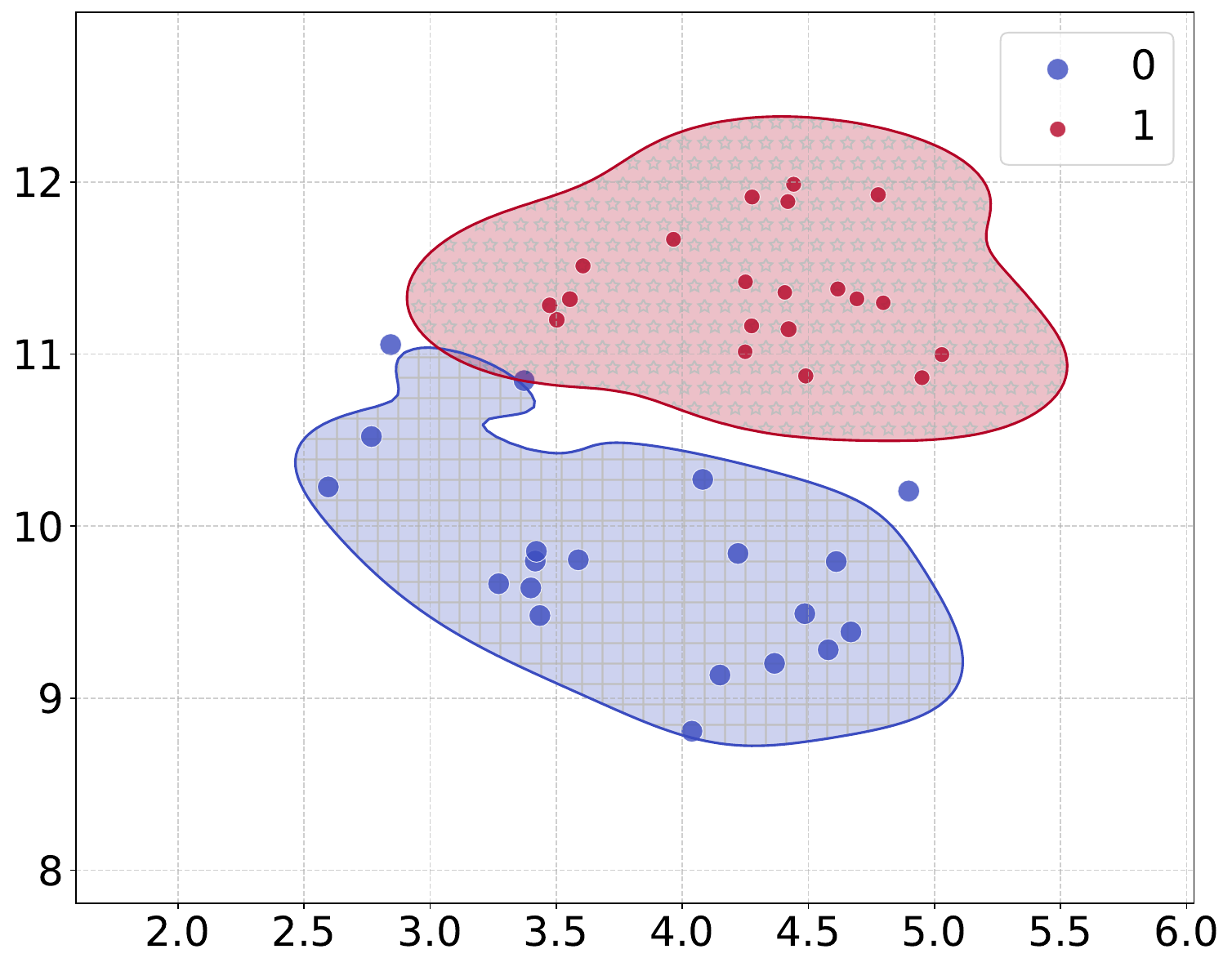}
}
\caption{t-SNE visualization of prototype sentence embeddings across five classification tasks. Each point represents a sentence, colored by its class label, while shaded regions show the kernel density estimate for each class. 
}
\label{fig:tsne}
\end{figure*}


To evaluate the structural coherence of prototype sentences $\mathcal{S}y^\star$ in embedding space, we measure intra-class compactness and inter-class separability. These properties reflect whether the selected prototypes form coherent, label-specific regions. We compute intra-class distance $d_{\text{intra}}$ and inter-class distance $d_{\text{inter}}$, defined in Equations~\ref{eq:d_intra_code} and~\ref{eq:d_inter_code}, respectively, and use their ratio $r = d_{\text{inter}} / d_{\text{intra}}$ as a class separability metric: $r > 1$ suggests meaningful label structure.

We conduct this analysis using SBERT~\cite{sentence_bert} and SimCSE~\cite{gao2021simcse}, two widely adopted sentence embedding models trained with contrastive learning, known for their effectiveness in capturing fine-grained semantic similarity. As shown in Table~\ref{tab:embedding-geometry}, for all text classifiers, the separability ratio $r$ remains consistently above 1.0. This indicates that the prototypes occupy well-defined regions in embedding space, with intra-class cohesion and inter-class distinction preserved. The highest values of $r$ are observed in the spam detection and jailbreak detection tasks—reaching up to 1.92 for SBERT, which suggests particularly strong inter-class separability in their latent representations. In contrast, sentiment classification yields the lowest separability ($r = 1.07$ for SBERT and $1.12$ for SimCSE), likely due to the ordinal and overlapping nature of sentiment labels ranging from 1 to 5.

\subsubsection{Visualization of Embedding Geometry.}


To qualitatively assess the structure of the recovered semantic distributions, we visualize the prototype sentence embeddings using t-SNE. As shown in Figure~\ref{fig:tsne}, each point represents a sentence embedding, with color denoting its predicted label and point size proportional to its estimated sampling radius $\alpha$.

Across all classification tasks, the visualizations reveal well-formed clustering patterns: samples within the same label tend to form compact groups, while inter-class regions are largely separable. Notably, the sentiment analysis task exhibits partial overlap between adjacent clusters, consistent with our earlier geometric analysis. This phenomenon reflects the ordinal structure of sentiment labels, which span a graded polarity continuum (e.g., from 1 to 5) rather than mutually exclusive categories. Such overlaps are expected, as semantically similar instances may reside near class boundaries in the embedding space.


\begin{table*}[t]
\centering
\small
\begin{tabular}{p{4.0cm} p{12cm}}
\toprule
\textbf{Label} & \textbf{Representative Constructed Samples} \\
\midrule

\textbf{0 (inferred negative semantics)} &
\parbox[t]{12cm}{
\textit{The company’s new policy is quite nasty to employees.}\\
\textit{The poet’s cryptical language made it challenging for readers to interpret his work.}\\
\textit{The politician’s ingratiatory remarks were met with widespread criticism from the public.}
}
\\

\midrule

\textbf{1 (inferred positive semantics)} &
\parbox[t]{12cm}{
\textit{The company will satisfy its customers with the new product launch.}\\
\textit{His perpetually optimistic attitude made him a joy to be around.}\\
\textit{After being cooked, the meal was surprisingly delicious.}
}
\\

\bottomrule
\end{tabular}
\caption{Constructed example sentences used to examine the semantic tendencies of each label.}
\label{tab:constructed-samples}
\end{table*}

\subsubsection{Effectiveness of Prompt-Free Sampler.}

We evaluate the quality of the prompt-free sampler by analyzing its outputs over the set of prototype sentences $\mathcal{S}_y^\star$. Specifically, we assess whether the generated samples preserve the semantics of the original sentence, maintain classification consistency under the target label, and exhibit sufficient diversity to support generalizable concept induction. To quantify semantic preservation, we compute the BERTScore between each seed sentence and its generated variants, capturing the degree of meaning retention. Classification consistency is measured as the percentage of generated sentences that are still classified as label $y$ by the black-box model $\mathcal{M}$, indicating whether the sampler remains within the decision boundary of the intended class. To assess lexical variability, the Self-BLEU (abbreviated as BLEU) is computed, indicating greater divergence from the original phrasing.

As shown in Table~\ref{tab:t5-generation-eval}, the prompt-free sampler consistently produces semantically faithful paraphrases $(\text{BERTScore} \ge 0.93)$ that retain the original label with perfect classification consistency across all tasks. Moreover, BLEU scores remain below 0.30, indicating that the generated samples avoid collapse and provide diverse instantiations of each concept. These results confirm the effectiveness of the sampler in constructing label-consistent, semantically rich distributions $\mathbb{D}_y$ that serve as a foundation for downstream analysis.

\subsection{Real Case Study: Forensics of an Undocumented HuggingFace Classifier}

To demonstrate the practicality of our label-forensics framework, we conduct a real-world case study on the HuggingFace model hub, one of the largest publicly accessible repositories of machine learning models. The platform hosts more than 106,000 text-classification models covering diverse nlp tasks.


Specifically, we select a newly uploaded binary text classifier\footnote{https://huggingface.co/sanderhs1/trained-mydata}
which contains no documentation, no dataset description, and no explanation of label meanings. Our goal is to construct an embedding representation for each output label and use it to infer both the task identity and the latent meaning of the labels. We begin our analysis by examining the semantic characteristics associated with each label without relying on any reference models, allowing us to observe how the classifier groups different types of inputs.

Using our label-forensics pipeline, we generate paraphrases for label~0 and label~1 and inspect their thematic patterns. As shown in Table~\ref{tab:constructed-samples}, the constructed samples display a clear and consistent contrast: sentences assigned to label~0 tend to carry negative or critical tones, while those assigned to label~1 generally express positive or supportive meanings. This pattern is stable across multiple paraphrases and remains visible even when we increase the diversity of the generated samples. Such consistency indicates that the classifier is not assigning these labels arbitrarily but is grouping inputs according to the broad sentiment expressed in the text. Therefore, we infer that label~0 and label~1 correspond to different sentiment directions, with one leaning negative and the other leaning positive.

To further verify the correctness of our forensic analysis, we compute similarity scores using cosine similarity and compare the target model’s label-specific paraphrase embeddings against all labels of each reference classifier. Table~\ref{tab:label_level_case} summarizes the closest semantic matches. Both labels of the target model align most strongly with the sentiment classifier, where target label~0 corresponds to \emph{Very Negative} and target label~1 corresponds to \emph{Very Positive}. This confirms the model-level attribution and provides a fine-grained interpretation of its label semantics.

\begin{table}[h]
\centering
\begin{tabular}{c|l|l|c}
\toprule
\textbf{Target Label} & \textbf{Reference Model} & \textbf{Closest Label} & \textbf{Similarity} \\
\midrule
\multirow{5}{*}{0}
& Emotion & Sadness & 0.1472 \\
& Sentiment & Very Negative & \textbf{0.3585} \\
& Spam & Ham & 0.3118 \\
& Toxicity & Non-Toxic & 0.2760 \\

& Jailbreak & Clean Prompt & 0.1559 \\

\midrule
\multirow{5}{*}{1}
& Emotion & Anger & 0.2779 \\
& Sentiment & Very Positive & \textbf{0.3158} \\
& Spam & Ham & 0.2357 \\
& Toxicity & Non-Toxic & 0.2332 \\

& Jailbreak & Jailbreak & 0.2513 \\

\bottomrule
\end{tabular}
\caption{Label-level alignment between the target classifier and reference models.}
\label{tab:label_level_case}
\end{table}

\section{Conclusion}

The rapid growth of web-scale text classifiers has made understanding their underlying label semantics increasingly important. In this paper, we introduce a label forensics framework that reconstructs the semantic behavior of a text classifier. Our framework consists of three components: a WordNet-based initialization that builds an anchor pool for each label, a prompt-free encoder–decoder sampler that expands these anchors into a label-specific sentence distribution, and a semantic interpretation module that derives coherent label meanings from the reconstructed distributions. Using this pipeline, our method enables investigators to reason about a classifier’s true behavior using only hard-label outputs. Experiments across five representative classifiers demonstrate that the recovered embedding regions achieve an average label consistency of 92.24\%, faithfully capturing decision-level semantics and exhibiting strong stability under sampling-based perturbations.


Beyond benchmark models, we further validate the framework on an undocumented HuggingFace classifier, showing that label forensics can reliably infer task identity and latent label meanings even in the absence of any documentation. This underscores the practical value of our approach for auditing real-world deployed models and supporting more transparent and trustworthy Web AI systems.



\bibliographystyle{ACM-Reference-Format}
\bibliography{refs}

\appendix

\end{document}